\pdfoutput=1

\documentclass[11pt]{article}

\usepackage[final]{coling}

\usepackage{times}
\usepackage{latexsym}

\usepackage[T1]{fontenc}

\usepackage[utf8]{inputenc}

\usepackage{microtype}

\usepackage{inconsolata}

\usepackage{graphicx}

\usepackage{booktabs}
\usepackage{tabularx}
\usepackage{caption}
\usepackage{amsfonts}
\usepackage{amsmath}
\usepackage{color,xcolor} 
\usepackage{xspace}
\usepackage{adjustbox}
\usepackage{multirow}

\newcommand*{\ie}{i.e.\@\xspace}

\usepackage{amssymb}
\usepackage{algpseudocode}
\usepackage{xcolor,colortbl}
\usepackage{algorithm}
\usepackage{dcolumn}

\title{JuniperLiu at CoMeDi Shared Task: Models as Annotators in Lexical Semantics Disagreements}

\author{
 Zhu Liu\textsuperscript{1}\thanks{These authors contributed equally.}, Zhen Hu\textsuperscript{2}\footnotemark[1], Ying Liu\textsuperscript{1} \\ 
 \textsuperscript{1}School of Humanities, Tsinghua University, Beijing, China\\
 \textsuperscript{2}College of Engineering, Beijing Forestry University, Beijing, China\\
 \texttt{liuzhu22@mails.tsinghua.edu.cn, huzhen@bjfu.edu.cn}
}

\begin{document}
\maketitle
\begin{abstract}

We present the results of our system for the CoMeDi Shared Task, which predicts majority votes (Subtask 1) and annotator disagreements (Subtask 2). Our approach combines model ensemble strategies with MLP-based and threshold-based methods trained on pretrained language models. Treating individual models as virtual annotators, we simulate the annotation process by designing aggregation measures that incorporate continuous relatedness scores and discrete classification labels to capture both majority and disagreement. Additionally, we employ anisotropy removal techniques to enhance performance. Experimental results demonstrate the effectiveness of our methods, particularly for Subtask 2. \textcolor{black}{Notably, we find that standard deviation on continuous relatedness scores among different model manipulations correlates with human disagreement annotations compared to metrics on aggregated discrete labels.} The code will be published at \url{https://github.com/RyanLiut/CoMeDi_Solution}.

\end{abstract}

\section{Introduction}
Lexical semantic similarity is a classical task that encompasses various forms, including multi-choice sense selection~\cite{2009wsd}, binary classification~\cite{pilehvar2019wic}, and contextual word similarity~\cite{islam2008similarity}, among others. 
However, the potential disagreements among annotators, arising from the inherent vagueness and continuous nature of meaning, have received comparatively less attention. To address these complexities, the CoMedi workshop (Context and Meaning - Navigating Disagreements in NLP Annotations\footnote{\url{https://comedinlp.github.io/}}) introduced a Shared Task with two subtasks~\cite{schlechtweg2025comedi}. 
Subtask 1 involves predicting the median judgment classification across four candidate labels, which represent the degree of similarity for a target word in context. Subtask 2 focuses on predicting annotator disagreement, which can be interpreted as a form of predictive uncertainty estimation~\cite{gal2017deep}.

In this paper, we first conceptualize the two subtasks as corresponding to two fundamental statistical properties of a Gaussian distribution: the mean and variance. Subsequently, we model each system, parameterized by specific variables, as an individual human annotator. These variables encompass both homogeneous factors, such as layers within the same model, and heterogeneous factors across different models. To address the tasks, we employ MLP-based and threshold-based approaches to generate continuous relatedness~\footnote{\textcolor{black}{We distinguish \textit{similarity} from \textit{relatedness}, with the task focusing on annotating relatedness scores.
}} scores and discrete classification labels, respectively. Additionally, we incorporate techniques for anisotropy removal to mitigate geometric biases inherent in embedding spaces. Finally, we propose diverse strategies for model ensembling to enhance performance. Our results demonstrate the effectiveness of threshold-based methods combined with anisotropy removal and MLP-based approaches. For Subtask 2, the findings further highlight the advantages of aggregating relatedness scores over discrete labels in capturing annotator disagreement.

\section{Related Work}

\paragraph{Probing for Contextual Word Meaning}  
Tasks capturing word meaning in context include word sense disambiguation (WSD)~\cite{2009wsd}, which selects the most appropriate sense, and WiC~\cite{pilehvar2019wic}, which determines semantic equivalence across contexts. Extending these, relatedness scoring provides a continuous measure of semantic relatedness. The CoMeDi Shared Task reframes WiC as an ordinal classification task with four labels indicating relatedness degrees. Probing methods include MLP-based approaches~\cite{tenney2019you,pilehvar2019wic}, which train dense networks, and threshold-based methods~\cite{pilehvar2019wic,vulic2020probing,liu-etal-2024-fantastic}, which optimize relatedness thresholds for pretrained representations. Since embeddings are often anisotropic~\cite{ethayarajh2019how}, techniques like centering~\cite{centering} and standardization~\cite{standardization} are applied to improve representation quality.

\paragraph{Uncertainty Estimation} Subtask 2 models annotator disagreement, aligning with the study of uncertainty estimation (UE), widely explored in computer vision~\cite{gal2017deep} and robust AI~\cite{stutz2022understanding}. UE arises from data uncertainty (aleatoric, linked to inherent data ambiguity like annotation disagreement) and model uncertainty (epistemic, due to biased learning on out-of-distribution data)~\cite{gal2017deep}. Researchers~\cite{liu2023ambiguity} combine these areas to model semantic uncertainty in sense selection. While Bayesian~\cite{vazhentsev2022uncertainty} and non-Bayesian~\cite{LabelSmooth} methods often use label probabilities, our threshold-based method lacks this feature. Instead, we treat the process as model ensemble~\cite{deepensemble} and propose aggregation measures.

\paragraph{Annotator Disagreement} 
Annotator disagreement is common in lexical semantics tasks, such as word sense disambiguation (WSD)~\cite{2009wsd,chklovski2003exploiting}, due to the subjective and ambiguous nature of meaning~\cite{navigli2008structural}. While many studies resolve disagreement through majority voting, others exploit it by reframing tasks as multi-label classification~\cite{conia2021framing} or training on multiple judgments~\cite{uma2021learning}. 

In this paper, we model annotator disagreement as uncertainty estimation, as both involve (1) output variability, (2) data noise~\footnote{Annotator disagreement can be viewed as label noise, contributing to data uncertainty—a key component of irreducible uncertainty.}, and (3) similar evaluation metrics.

\section{System Overview}

Most systems use MLP-based~\cite{tenney2019you,pilehvar2019wic} and threshold-based~\cite{pilehvar2019wic,vulic2020probing,liu-etal-2024-fantastic} methods. They extract representations from pretrained language models, then MLP-based methods train a network to predict discrete labels (Subtask 1) or continuous values (Subtask 2). Threshold-based methods learn a threshold selector to map similarity scores to labels. However, naive baselines often fall short, as shown in Section~\ref{sec:results}. In our system, we applied anisotropy removal to the baseline code~\cite{schlechtweg2025comedi} and used a classifier-based method for comparison. For Subtask 1, we apply techniques to make data points more isotropic. For Subtask 2, we ensemble models, treating them as annotators, and use various strategies to model disagreement.

\subsection{Formulation as Parameter Estimation}

For a target word \(w\) appearing in a pair of contexts \(c_i\) and \(c_j\), annotators from a hypothetical human space \(\mathcal{H}\) provide a judgment score \(s \in \mathcal{R}\), where higher values indicate greater similarity in meaning between \(c_i\) and \(c_j\). These scores form a judgment distribution \(p\) on \(\mathcal{R}\), which we assume follows a Gaussian distribution, \(p \sim \mathcal{N}(\mu, \sigma^2)\), as it is a natural statistical choice~\cite{jaynes2003probability}. Here, \(\mu\) represents the consensus similarity, while \(\sigma\) reflects disagreement among annotators.

In practice, the continuous Gaussian distribution is discretized due to the finite number of annotators and graded annotations. Nonetheless, we adopt the Gaussian framework to unify the two tasks: Subtask 1 estimates \(\mu\), while Subtask 2 estimates \(\sigma\).


\subsection{Subtask 1: Anisotropy Removal}
\textcolor{black}{Contextual representations are known to be anisotropic~\cite{ethayarajh2019how}, clustering in a narrow region of the space. This inflates similarity scores, reducing their discriminative power in meaning-related tasks. For example, even unrelated words often exhibit high similarity. We adopt three techniques to reduce anisotropy:} (1) centering by subtracting the mean vector (2) normal standardization (3) All-but-the-top~\cite{all-but-the-top}: subtracting the projection on the component of the largest variance.


\subsection{Subtask 2: Model Ensembling}
\label{subsection2}
To model annotator disagreement, we treat each model or its manipulation as an annotator and use three measures to reflect uncertainty. We explore three ensembling strategies: (1) homogeneous aggregation with model manipulations (e.g., layer and anisotropy removal), (2) heterogeneous ensembling across different models, and (3) a mixed approach combining both. After each model forward pass, we obtain a discrete label using the threshold-based model~\footnote{For Subtask 2, we use the majority of judgment scores as the GT label, avoiding the median to handle decimals.} and a continuous relatedness score. We apply three measures: standard deviation (STD) for continuous scores, mean pairwise absolute judgment differences (MPD) for discrete labels (as used in Subtask 2), and variation ratio (VR), the ratio of values not equal to the mode, commonly used in uncertainty estimation~\cite{gal2017deep}.

\section{Experiment Setup}

\begin{table*}[t]
    \centering
    \small
    \begin{tabular}{cc|cccccccc}
    \toprule
        Participator & Method & AVG & ZH & EN & DE & NO & RU & ES & SV \\
        \midrule
        kuklinmike & - & \textbf{0.656} & \textbf{0.424} & \textbf{0.732} & 0.723 & \textbf{0.668} & \textbf{0.623} & \textbf{0.748} & \textbf{0.675} \\
        comedy\_baseline\_2 & - & 0.583 & 0.379 & 0.654 & \textbf{0.728} & 0.515 & 0.550 & 0.656 & 0.601 \\
        daalft & - & 0.555 & 0.317 & 0.555 & 0.656 & 0.589 & 0.487 & 0.636 & 0.648 \\
        ours & Thr* (XLM-R-B) & 0.271 & 0.140 & 0.507 & 0.492 & 0.080 & 0.128 & 0.330 & 0.224 \\
        \midrule
        ours & Thr (LLM) & \textbf{0.451} & -0.090 & 0.474 &\textbf{ 0.696} & \textbf{0.445} & \textbf{0.444} & \textbf{0.623} & \textbf{0.566} \\

        ours & Thr (XLM-R-B) & 0.339 & \textbf{0.148} & \textbf{0.524} & 0.485 & 0.240 & 0.301 & 0.348 & 0.325 \\
        
        ours & MLP & 0.338 & 0.128 & 0.369 & 0.371 & 0.351 & 0.329 & 0.411 & 0.407 \\
    \bottomrule
    \end{tabular}
    \caption{Results for Subtask 1. The upper part shows the evaluation phase, and the lower part the post-evaluation phase. ``Thr'' denotes threshold-based methods, and ``Thr*'' indicates language-specific model selections. The same applies to other tables.}

    \label{tab:sub1}
\end{table*}

\begin{table*}[t]
    \centering
    \small
    \begin{tabular}{cc|cccccccc}
    \toprule
        Participator & Method & AVG & ZH & EN & DE & NO & RU & ES & SV \\
        \midrule
        kuklinmike & - & \textbf{0.226} & 0.301 & 0.078 & 0.204 & \textbf{0.286} & \textbf{0.175} & \textbf{0.187} & 0.350 \\
        daalft & - & \underline{0.220} & \textbf{0.539} & 0.042 & 0.108 & 0.272 & \underline{0.167} & \underline{0.115} & 0.296 \\
        comedy\_baseline\_2 & - & 0.163 & \underline{0.485} & 0.060 & 0.085 & 0.235 & 0.116 & 0.078 & 0.079 \\
        ours & MLP & 0.082 & 0.358 & 0.038 & 0.022 & -0.042 & 0.067 & 0.040 & 0.090 \\
        \midrule
        ours & ensembling & 0.205 & 0.274 & \underline{0.117} & \underline{0.236} & 0.279 & 0.101 & 0.073 & \underline{0.353}  \\
        ours & ensembling* & \underline{0.220}  & 0.347 & \textbf{0.118} & \textbf{0.242} & \underline{0.283} & 0.108 & 0.078 & \textbf{0.364} \\
    \bottomrule
    \end{tabular}
    \caption{Evaluation results (upper part) and post-evaluation results (lower part) for Subtask 2. The method \textit{ensembling*} integrates language-specific ensembling strategies, while \textit{ensembling} uses the strategy with the best average score across all languages.}
    \label{tab:sub2}
\end{table*}

\subsection{Task Description}
The Shared Task in the workshop of CoMeDi~\cite{schlechtweg2025comedi} includes two subtasks. The first aims to predict a discrete label (from 1 to 4) to show the relatedness of the target word in two contexts while the second obtains a continuous value to indicate the disagreement. The task data was sampled from multilingual datasets, involving 7 languages, i.e., Chinese~\cite{Chen2023chiwug}, English~\cite{schlechtweg-etal-2021-dwug, Schlechtweg2024dwugs}, German~\cite{schlechtweg-etal-2018-diachronic,schlechtweg-etal-2021-dwug,  Schlechtweg2024dwugs, hatty-etal-2019-surel, Kurtyigit2021discovery, DBLP:phd/dnb/Schlechtweg23}, Norwegian~\cite{kutuzov2022nordiachange}, Russian~\cite{rodina2020rusemshift, rushifteval2021, Aksenova2022rudsi}, Spanish~\cite{Zamora2022lscd}, Swedish~\cite{schlechtweg-etal-2021-dwug, Schlechtweg2024dwugs}. 

\subsection{Models}
Our study focuses on threshold-based methods using pre-trained models: XLM-RoBERTa-base, XLM-RoBERTa-large~\cite{conneau2019unsupervised}, BERT-multi-base~\cite{pires2019multilingual}, and Llama-7B~\cite{touvron2023llama}. For encoder-only models, we extract target word representations directly, while for the decoder-only Llama-7B, we use a prompt-based method~\cite{liu2023ambiguity} to extract the final colon representation. Inspired by in-context learning~\cite{jiang-etal-2024-scaling}, we apply layer-wise manipulations (centering, standardization, and all-but-the-top) to reduce anisotropy. 

We discretize the continuous similarity scores into labels using a threshold selector based on the shared task baseline. The selector employs the Nelder-Mead method~\cite{nelder1965simplex} to optimize bin edges for Krippendorff's $\alpha$, starting with evenly spaced bins and iteratively refining them. For Subtask 2, we explore model ensembling strategies (homo, hetero, mixed) and different measures (STD, MPD, VR), and also evaluate an MLP-based approach (details in Appendix~\ref{app:setting}).

\subsection{Evaluation Phase Setting}
During the evaluation phase, we selected models based on the development set.  

For \textbf{Subtask 1}, we employ a threshold-based method using XLM-RoBERTa-base as the pretrained model, except for Chinese and Russian (BERT-multi-base) and Norwegian (LERT-base-chinese). Representations are extracted from the 10th layer for XLM-RoBERTa-base and the final layer for other models. We apply normal standardization except for Norwegian to address anisotropy and utilized the threshold selection method from the official baseline code~\cite{schlechtweg2025comedi}. Specifically,   

For \textbf{Subtask 2}, we fine-tune an MLP regressor to predict disagreement scores, following the baseline methodology. It comprised of two linear layers and a ReLU activation function. For Swiss, we train for 50 epochs with a batch size of 32, while for other languages, we use 200 epochs with a batch size of 16. The learning rate is 1e-2 with a 0.1 dropout rate. We utilize AdamW for optimization with a warm-up ratio of 0.1. 

\subsection{Post Evaluation Phase Setting}
For \textbf{Subtask 1}, we use the 25th layer of Llama and the 11th layer of XLM-RoBERTa-Base for all languages, with an MLP-based method fine-tuned using training data. All model representations are standardized to remove anisotropy. For the MLP-based model, we train for 50 epochs with a batch size of 128, an initial learning rate of 1e-2, and apply a dropout rate of 0.1 to prevent overfitting.

For \textbf{Subtask 2}, we employ ensembling strategies to significantly improve performance. We report two results from our ensembling methods. The first (ensembling) applies the same strategy across all languages: standardization with layer 24, no standardization with layer 16, centering with layer 24, and all-but-the-top with layer 16, all on Llama-7B. The second (ensembling*) presents language-specific ensembling strategies, as in Table~\ref{tab:specific_lanuage}.



\section{Results}
\label{sec:results}


We present the results in Table~\ref{tab:sub1} and Table~\ref{tab:sub2} on the \textbf{test} set. The upper sections show evaluation phase scores submitted to the leaderboard, while the lower sections display post-evaluation results using public answers. We then conduct abalation studies on the \textbf{development} set in later sections.  

In the evaluation phrase, for \textbf{Subtask 1}, our threshold-based method achieved moderate results, with LERT-base-chinese performing relatively better for Norwegian, though with limitations. For \textbf{Subtask 2}, we fine-tuned an MLP to predict disagreement scores but observed limited performance, prompting alternative methods in the post-evaluation phase.  

In the post-evaluation phase, for \textbf{Subtask 1}, the threshold-based model performed comparably to the MLP-based model, while large language models (LLMs) showed superior results, highlighting their potential. For \textbf{Subtask 2}, our results matched the evaluation phase’s top performances, confirming the effectiveness of the ensembling approach.


\subsection{Abalation Study on Subtask 1}

\begin{figure}
    \centering
    \includegraphics[width=0.9\linewidth]{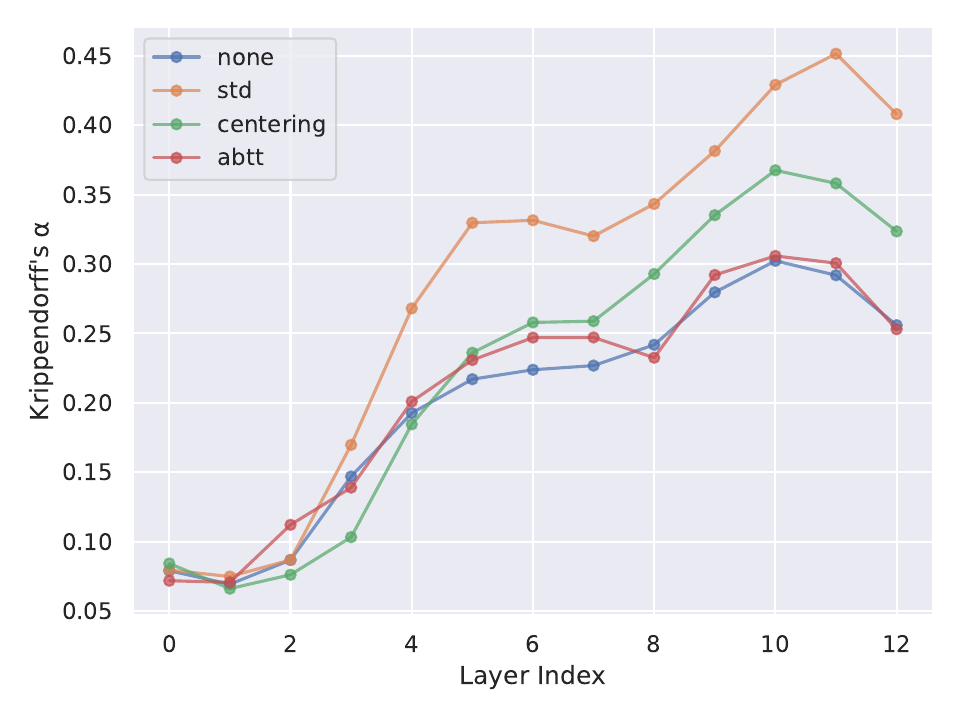}
    \caption{Performance of different types of anisotropy removal with the increase of layer index. 0 indicates the input embedding. ``abtt'' means all-but-the-top.}
    \label{fig:stds}
\end{figure}

Figure~\ref{fig:stds} shows the average performance change with different anisotropy removal methods across layers. The large gap between removal and non-removal emphasizes the importance of this technique. Performance improves with higher layers, except for a drop in the last one or two layers. Standardization consistently performs best across all layers.

\begin{figure}
    \centering
    \includegraphics[width=0.9\linewidth]{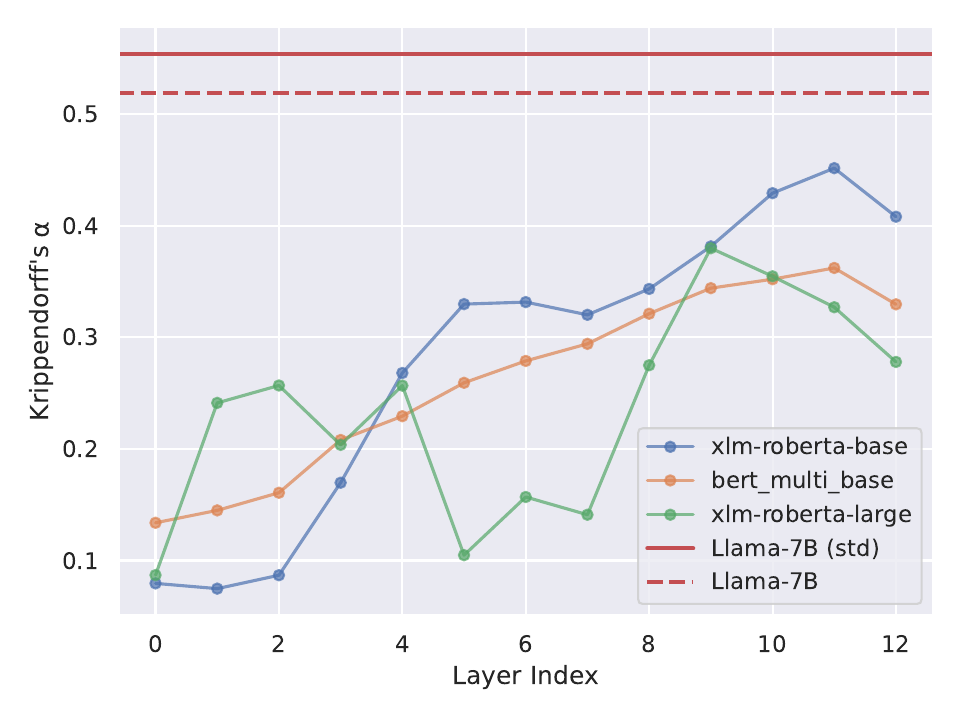}
    \caption{Performance of different models as the layer index increases. The optimal result (Layer 25) for Llama-7B and its standardized version are shown as the upper bound.}
    \label{fig:models}
\end{figure}

Figure~\ref{fig:models} displays the performance of different models. Since Llama-7B is a decoder-only model with significantly more parameters and training data, its optimal result (Layer 25) serves as an upper bound~\footnote{We attempt representations of different layers from Llama-7B, and the optimal layer index is 25.}. The results show that XLM-RoBERTa-base outperforms all other models, including its larger counterpart.

\subsection{Abalation Study on Subtask 2}
In this section, we analyze various factors influencing ensembling performance, including the choice of measure and model selection. We evaluate four candidate models \ie, XLM-RoBERTa-base, XLM-RoBERTa-large, BERT-multi-base, and Llama-7B, with four types of anisotropy removal and four layer levels. For layer levels, we extract layers 1, 4, 7, 10 for encoder-only models, and layers 8, 16, 24, 32 for the Llama model, yielding 64 possible model configurations. We use a threshold-based method for each model to obtain both a continuous similarity score and a discrete classification label, as we have done in Subtask 1. We randomly select a subset of 4 models from these possibilities, referred to as "mixed". Additionally, we experiment with homogeneous aggregation (using the same model) and heterogeneous aggregation (using different models). For homogeneous aggregation, we choose Llama-7B due to its superior performance. For each category, we sample 500 model subsets, obtaining both their classification labels using a threshold-based method and relatedness scores based on pre-trained embeddings. We first evaluate three measures (STD, MPD, and VR) in the mixed setting, selecting the best one to compare different category choices.

\begin{figure}
    \centering
    \includegraphics[width=1.0\linewidth]{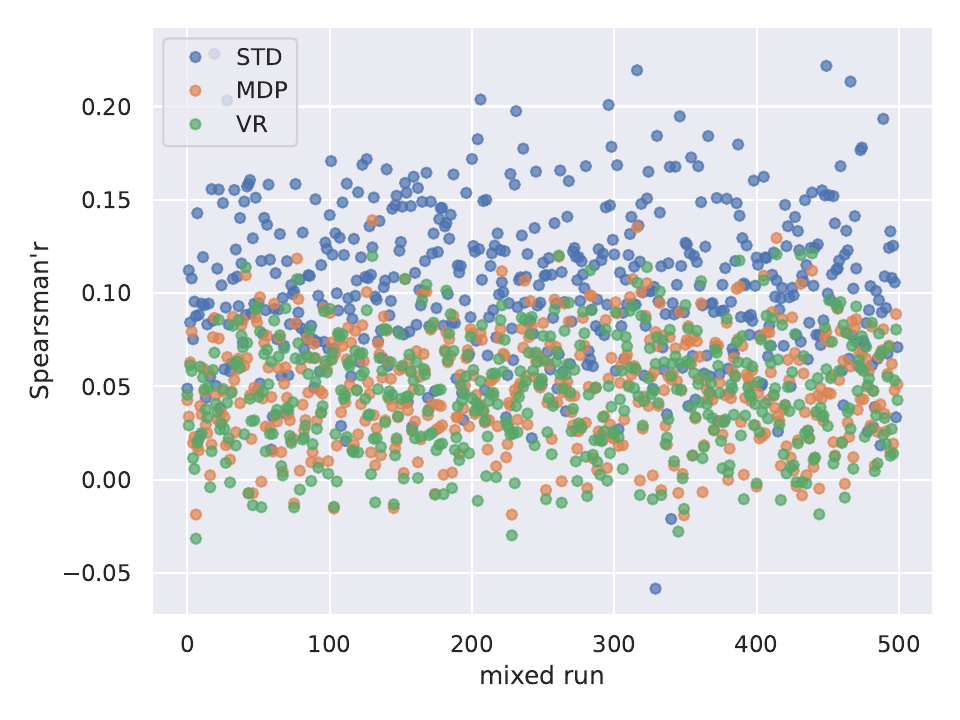}
    \caption{Performance of three types of measures across 500 random runs.}
    \label{fig:measure}
\end{figure}

\paragraph{Measure} 
Figure~\ref{fig:measure} presents the results for three measures. In most cases, STD on a continuous similarity score outperforms the others, while MDP slightly exceeds VR on the discrete classification labels . This suggests that similarity scores have an advantage over discrete labels due to the robustness of continuous values. Label prediction can be seen as a discretization of the continuous counterpart, leading to a loss of precision. Thus, we select STD as our final measure.

\paragraph{Model Selection} 
Table~\ref{tab:ensembles_type} shows the top five results for three ensemble strategies. The specific model groups are listed in Tabel~\ref{tab:ensembles_type_names}. Homogeneous model manipulations (homo) outperform mixed ensembles, while combining different models yields the worst performance. This suggests that model variance can still serve as an effective alternative, aligning with the use of dropout in uncertainty estimation~\cite{gal2017deep}.

\begin{table}[]
    \centering
    \small
    \begin{tabular}{cccccc}
    \toprule
        Type & 1 & 2 & 3 & 4 & 5 \\
    \midrule
        homo &\textbf{ 0.237 }&\textbf{ 0.235 }& \textbf{0.235} & \textbf{0.234} & \textbf{0.233} \\
        hete & 0.217 & 0.216 & 0.215 & 0.209 & 0.201 \\
        mixed & 0.228 & 0.222 & 0.219 & 0.213 & 0.203 \\
    \bottomrule
    \end{tabular}
    \caption{Top five groups for strategies of model selections}
    \label{tab:ensembles_type}
\end{table}


\section{Conclusion}
We present our system for two subtasks released on CoMeDi Shared Task. We first formalize these tasks as parameter estimation where Subtask 1 estimates a mean and Subtask 2 the variance for a hypothetical Gaussian distribution. Then we mainly adopt threshold-based method with different techniques of anisotropy removal to classify the label for Subtask 1. Inspired by the area of uncertainty estimation, we utilize model ensembling with various strategies to select models and measures to reflect disagreement for Subtask 2. Experiments show the effectiveness of our method. 

\section{Limitations}
We acknowledge several limitations in our system. First, the model training process utilizes data from all languages without considering their unique linguistic characteristics. For instance, Chinese exhibits rich formation rules~\cite{zheng-etal-2021-leveraging-word}, yet lacks the morphological complexity found in Western languages, potentially leading to distinct patterns of disagreement. Second, our parameter estimation for the Gaussian distribution does not account for the estimation of the mean, which could be incorporated into Subtask 1 for a more comprehensive approach. Furthermore, in Subtask 2, we employ the median of all annotations as an independent label for the model instead of using individual annotations. This approach may introduce inconsistencies with our formulation of \textit{models as annotators}. Lastly, while our experiments highlight the potential of large language models (LLMs) compared to pretrained language models, future work will focus on exploring more effective strategies for extracting lexical representations from LLMs.

\section{Ethics Statement}
We do not foresee any immediate negative ethical consequences arising from our research.

\section{Acknowledgements}
The authors thank the anonymous reviewers for their valuable comments and constructive feedback on the manuscript. This work is supported by the 2018 National Major Program of Philosophy and Social Science Fund “Analyses and Researches of Classic Texts of Classical Literature Based on Big Data Technology” (18ZDA238) and Research on the Long-Term Goals and Development Plan for National Language and Script Work by 2035 (ZDA145-6). 

\bibliography{custom}

\clearpage
\begin{table*}[t]
    \centering
    \small
    \begin{tabular}{cc|cccccccc}
    \toprule
        Method & AVG & ZH & EN & DE & NO & RU & ES & SV \\
        \midrule
         MLP1 & 0.191 & 0.105 & -0.140 & 0.192 & 0.337 & 0.276 & 0.418 & 0.151 \\
         weighted loss & 0.240 & 0.361 & 0.110 & 0.166 & 0.156 & 0.255 & 0.354 & 0.277 \\
         layer11 & 0.265 & 0.267 & 0.009 & 0.261 & 0.357 & 0.298 & 0.341 & 0.321 \\
        \midrule 
        MLP2 & 0.407 & 0.519 & 0.268 & \textbf{0.609} & 0.360 & 0.265 & 0.565 & 0.262 \\
        layer11 & 0.418 & 0.530 & \textbf{0.384} & 0.511 & \textbf{0.416} & 0.311 & 0.576 & 0.198 \\
        last4layer & \textbf{0.429} & \textbf{0.509} & 0.229 & 0.570 & 0.306 & \textbf{0.416} & \textbf{0.584} & \textbf{0.386} \\
    \bottomrule
    \end{tabular}
    \caption{Evaluation results for Subtask 1 in MLP-based methods. The upper part presents the outcomes of using a single linear layer as a classifier, where ``weight loss'' indicates the employment of a weighted cross-entropy loss function, and ``layer11'' denotes utilizing the vector representations from the 11th layer of the language model. The lower part illustrates the results obtained by employing two linear layers as classifiers, showing the performance of the 11th layer of the model as well as the outcome after applying average pooling to the last four layers of the model.}
    \label{tab:sub4}
\end{table*}

\begin{table*}[t]
    \centering
    \small
    \begin{tabular}{cc|cccccccc}
    \toprule
        Method & AVG & ZH & EN & DE & NO & RU & ES & SV \\
        \midrule
         MLP1 & \textbf{0.128} & \textbf{0.323} & \textbf{0.088} & \textbf{0.179} & \textbf{0.132} & \textbf{0.061} & 0.026 & 0.083 \\
         MLP2 & 0.098 & 0.232 & -0.061 & 0.131 & 0.119 & 0.020 & \textbf{0.061} & \textbf{0.187} \\
    \bottomrule
    \end{tabular}
    \caption{Evaluation results for Subtask 2 in MLP-based methods, demonstrating the results of Multi-Layer Perceptrons (MLPs) with different numbers of layers.}
    \label{tab:sub5}
\end{table*}

\begin{table*}[htbp]
    \centering
    \small
    \begin{tabular}{cccccc}
    \toprule
        Language & Model Groups \\
    \midrule
        Chinese & AiX-AkX-AhX-AkW  \\
        English & AjZ-AiX-AjX-AjW \\
        German & AhW-AjX-AjW-AjZ \\
        Norwegian & AjZ-AiX-AjX-AjW \\
        Russian & AiX-AiW-AkW-AkZ \\
        Spanish & AhY-AiZ-AhX-AhW \\
        Swedish & AiX-AkY-AjZ-AjY \\
    \bottomrule
    \end{tabular}
    \caption{The optimal model groups for each specific language for the development set in Subtask 2.}
    \label{tab:specific_lanuage}
\end{table*}

\begin{table*}[htbp]
    \centering
    \small
    \begin{tabular}{cccccc}
    \toprule
        Type & 1 & 2 & 3 & 4 & 5 \\
    \midrule
        homo &\textbf{ AjY-AjZ-AiX-AiW }&\textbf{ AjY-AiW-AjZ-AjX }& \textbf{AjX-AiX-AiW-AjY} & \textbf{AjZ-AiX-AjX-AjW} & \textbf{AiX-AjZ-AjX-AhX} \\
        hete & AjY-BkW-ChZ-DkX & AjY-BiX-ChW-DjX & AjY-BkW-CiX-DiW & AjZ-BkW-ChY-DhW & AjY-BhW-ChY-DiX \\
        mixed & AjX-ChX-AiX-AjZ & AjY-AiX-AjW-ChY & AkW-ChX-AjY-AjW & AjZ-ChY-DhX-DkX & ChX-AkY-AiX-AiW \\
    \bottomrule
    \end{tabular}
    \caption{Top five model groups when ensembling models for Subtask 2.}
    \label{tab:ensembles_type_names}
\end{table*}

\section{Appendix}

\subsection{MLP-based Methods}
\label{app:setting}
We attempt the MLP-based method in two subtasks, freezing XLM-RoBERTa-base model parameters to obtain the vector representation of the target word, and training a classifier or a regression model downstream. MLP1 is a linear layer, MLP2 represents two linear layers, and uses the ReLU activation function.

In Subtask 1, we use the cross-entropy loss function to train a classifier. The results on the development dataset are shown in Table~\ref{tab:sub4}. We find that two linear layers achieve better results. We attempt to use a weighted cross-entropy loss function to alleviate the problem of sample imbalance, but shows slight improvement. We compare the results of different layers of the model and find that the vector representation of the shallower layers(11th) achieves better results. We attempt layer fusion and average pooling of the vectors in the last 4 layers, which results in more stable improvements.

Training settings for the MLP-based method in Subtask 1: 50 epochs, batch size of 128, 1e-2 learning rate, AdamW optimizer, and a dropout rate of 0.1 to improve generalization.

In Subtask 2, we use the mean square error loss function to train a regression model that directly predicts continuous values of inconsistent labeling of target words, similar to the baseline provided by the official source. The results on the development dataset are shown in Table~\ref{tab:sub5}. We find that two linear layers are worse than a single linear layer.

We try multiple different hyperparameter settings on Task 2. In MLP1, we ultimately chose 200 epochs, batch size of 16, while in MLP2, we chose 50 epochs, batch size of 32. Other training settings: 1e-2 learning rate, AdamW optimizer, and a dropout rate of 0.1.

\subsection{Model Groups}
\label{app:models_specific_language and app:models}
We use letters to denote different models: A, B, C, and D represent Llama-7B, XLM-RoBERTa-base, BERT-multi-base, and XLM-RoBERTa-large, respectively. 

For encoder-only models, h, i, j, and k indicate layers 1, 4, 7, and 10, respectively; whereas in large language models (LLMs), these symbols correspond to layers 8, 16, 24, and 32. 

X, Y, Z, and W correspond to four standardization methods: non-standard, std, centering, and all-but-the-top.

\paragraph{Model groups for specific languages.}
We experiment with various model groups, and different groups achieve the best results in different languages. Table~\ref{tab:sub2} shows the best results for the test dataset in Subtask 2, and the specific model groups are shown in Table~\ref{tab:specific_lanuage}.

\paragraph{Top 5 model groups.}
We employ three ensemble strategies, and the top five results of each strategy on the development dataset of Subtask 2 are presented in Table 3, with corresponding model groups shown in Table~\ref{tab:ensembles_type_names}.

\end{document}